\documentclass{article}

\PassOptionsToPackage{numbers, compress}{natbib}

\usepackage[dblblindworkshop, final]{neurips_2026}

\usepackage[utf8]{inputenc} 
\usepackage[T1]{fontenc} 
\usepackage{hyperref} 
\usepackage{url} 
\usepackage{booktabs} 
\usepackage{amsmath}
\usepackage{amssymb}
\usepackage{amsfonts} 
\usepackage{nicefrac} 
\usepackage{microtype} 
\usepackage{xcolor} 
\usepackage{graphicx}
\usepackage{subcaption}
\usepackage{tabularx}

\title{Surface Conditioning for Zero-Shot Multivariate Correction in a Marine Biogeochemistry Emulator}
\workshoptitle{Sim2Science: ML with Imperfect Scientific Models}

\author{Gabriela Martínez Balbontín \\
Mercator Océan International \\
Sorbonne Université, LIP6, CNRS \\
\texttt{gmartinezbalbontin@mercator-ocean.fr} \\
\And Anastase Charantonis \\
INRIA Paris (ARCHES)\\
\And Dominique Béréziat \\
Sorbonne Université, LIP6, CNRS \\
\And Stefano Ciavatta \\
Mercator Océan International \\
}

\begin{document}
\maketitle

\begin{abstract}
  Bias correction in process-based models usually means iterative data assimilation with explicit covariance modelling. We propose a lightweight alternative based on a neural emulator of marine biogeochemistry (BGC), showing that the conditioning pathway of its autoencoder can serve as an observation injection point. The decoder is conditioned on surface variables via Feature-wise Linear Modulation (FiLM), originally designed to incorporate physical boundary forcings. We extend these inputs to include surface chlorophyll-a and replace the simulated field with satellite observations at inference time. The surface substitution propagates through the water column, improving agreement with depth-resolved observations at all depths, and spreads to other output variables in a pattern consistent with expected trophic relationships.
\end{abstract}

\section{Introduction}

Marine biogeochemical (BGC) models are essential for understanding ocean ecosystems and carbon-climate feedbacks~\cite{linkWhyWeNeed2023, park_seasonal_2019, gruberOceanicSinkAnthropic2019, moraBioticHumanVulnerability2013}, but they exhibit substantial biases relative to observations due to imperfect process representations that compound over time~\cite{fennel_ocean_2022, seferianTrackingImprovementSimulated2020, terhaarAssessmentGlobalOceanBiogeochemistry2024, fuEvaluationOceanBiogeochemistryCMIP2022}. The standard operational approach is data assimilation, where the simulation is periodically corrected by combining it with observations to create an \textit{analysis} before propagating the dynamical model on the improved state. This is typically done with variational or ensemble-based methods~\cite{bannister2017,Dowd2014,fennel_ocean_2022}, but it is computationally expensive, and multivariate corrections depend on error covariances that must be modelled, estimated, or inflated.

Machine learning has been applied to this problem along several axes, including learning direct mappings from model state to observation-space~\cite{fabletLearningVariationalData2021, liangMachineLearningApproach2023}, performing assimilation in a compressed latent space~\cite{peyronLatentSpaceData2021, fanPhysicallyConsistentGlobal2025}, replacing the dynamical model with a neural surrogate~\cite{brajardCombiningDataAssimilation2020, adrianDataAssimilationMachine2025}, and score-based data assimilation (SDA)~\cite{rozetScorebasedDataAssimilation2023}, to name a few. Most of these still pay an iterative cost per analysis, and require specifications of how information moves from observed to unobserved coordinates.

Instead we attempt something much simpler. We build on BG4Sea~\cite{BG4Sea}, a forecasting system trained on mechanistic model output, in which an autoencoder compresses the water column into a latent representation that a latent forecaster propagates forward in time. The decoder is conditioned on surface inputs via Feature-wise Linear Modulation (FiLM)~\cite{perezFiLMVisualReasoning2017}, originally designed to incorporate physical boundary forcing for the BGC forecasts. We repurpose this conditioning channel for observation injection, adding surface chlorophyll-a (chl-a) to the conditioning inputs during training and replacing it with observations at inference. Because this mechanism is decoupled from the latent representation and from the forecaster, this zero-shot replacement can propagate a depth-resolved correction through the decoder's learned prior, without retraining or a gain matrix. This is conceptually similar to adaptations of SDA~\cite{manshausenGenerativeDataAssimilation2025, martinGenerativeDataAssimilation2025}, which also correct the state at inference time without running a dynamical model forward. Our approach trades their flexibility and uncertainty estimates for a single forward pass.

Chl-a is a natural first choice for the ``assimilated'' variable. It is the pigment of the phytoplankton responsible for about half of Earth's photosynthesis~\cite{field_primary_1998}, and an indicator of primary production and carbon export. It colours the ocean enough to be seen from space, making it the best-observed BGC variable and the one most commonly assimilated into BGC models~\cite{ford_assimilating_2012, fennel_ocean_2022}, and since it varies with depth, it is also profiled in situ by autonomous floats. This dual coverage shapes our setup: we inject satellite observations at the surface and validate against a depth-resolved float product. The two observation sources disagree significantly, both with each other and with the model prior, so we also report their mutual agreement as a reference for our evaluation. Our results show that surface conditioning can improve accuracy at all depths with a trophically consistent response across other output variables. The surface conditioning alone shows similar results to full-column replacement with the target in an unmodulated network. In addition, the analysis agrees with the float-based product more closely than the injected satellite observation itself.

\section{Methodology}

For this study, we focus solely on the autoencoder, leaving the dynamical system fixed, the underlying assumption being that a perfect version of the latent forecast would coincide with the autoencoder's latent representation of the same time period.

\paragraph{Data}

The autoencoder is trained on normalised anomaly ocean data from mechanistic model output, both produced by Mercator Océan International and distributed by the Copernicus Marine Environment Monitoring Service. These include both BGC (BIORYS4~\cite{gregoire_review_2026, lamouroux_global_2023}) and physical (GLORYS12~\cite{lellouche_copernicus_2021, noauthor_global_nodate-3}) variables. Anomalies are calculated by subtracting a climatology (i.e. historical mean) from the data, and subsequently normalised using mean and standard deviation. Both are monthly at 1/4° horizontal resolution and split into training (2010--2017), validation (2018--2019) and test (2020--2022) periods. The injected chl-a observations come from GlobColour, a cloud-free product merging several satellite sensors~\cite{garnesson_cmems_2019-1, noauthor_global_nodate-5}. For validation we use SOCA~\cite{noauthor_global_nodate-4}, which interpolates profiles from BGC-Argo floats (autonomous instruments that measure chl-a down the water column) onto a regular grid~\cite{Renosh2023,sauzedeNeuralNetworkBasedMethod2016}. Despite targeting the same variable, the two observational sources differ significantly. We attempt to overcome these discrepancies by converting the column profiles to a satellite-equivalent surface value following standard practice~\cite{gordonRemoteSensingOptical1980, morelSurfacePigmentsAlgal1989, morelExaminingConsistencyProducts2007}, but the disagreement persists. Satellite chl-a is known to have a narrower range than float measurements and thus underestimates high concentrations~\cite{huChlorophyllAlgorithmsOligotrophic2012, baudenaComparingSatelliteBGCArgo2025, johnsonThreeImprovedSatellite2013, garnesson_cmems_2019-1}. We detrend and normalise both datasets using the same climatology and statistics used for the emulator's inputs.

\paragraph{Architecture}

The autoencoder is a deterministic, 1D convolutional network that compresses each water-column profile $\mathbf{x} \in \mathbb{R}^{V \times K}$ ($V = 31$ variables, $K = 24$ depth levels; Tables~\ref{tab:all_vars}--\ref{tab:depths}) into a latent code $\mathbf{z} = \mathcal{E}(\mathbf{x}) \in \mathbb{R}^{L}$, where $L=32$. The decoder mirrors the encoder, except that each stage is modulated via FiLM, which scales and shifts its feature maps according to a conditioning input. This input is a 128-dimensional vector produced by a two-layer MLP from the surface variables $\mathbf{s}$\footnote{Mixed-layer depth, sea-surface height, salinity and temperature, particulate dust, and, in this work, chl-a (Table~\ref{tab:surface_cond}).} and a learned embedding of month, latitude, longitude and depth. The original BG4Sea network is conditioned on physical variables only; the adapted autoencoder has identical architecture but extends the conditioning to $\mathbf{s} = [\mathbf{s}_{\mathrm{phys}}, s_{\mathrm{chl}}]$, and is trained to reconstruct $\mathbf{x}$ as $\mathcal{D}\big(\mathbf{z};\, \mathbf{s}_{\mathrm{phys}}, s_{\mathrm{chl}}^{\mathrm{sim}}\big)$. At inference, replacing the simulated surface chl-a with its GlobColour observation yields the analysis $\hat{\mathbf{x}}^{a} = \mathcal{D}\big(\mathbf{z};\, \mathbf{s}_{\mathrm{phys}}, s_{\mathrm{chl}}^{\mathrm{obs}}\big)$.

The analysis is compared to the emulator baseline, and with BIORYS4, and all are evaluated against SOCA. To isolate the contribution of the conditioning pathway from the variable replacement, two diagnostic runs feed the validation target SOCA directly: an oracle run replaces both the column and surface chl-a with SOCA, giving an upper bound, while an ablation replaces only the column chl-a in a column-only autoencoder (same architecture without the FiLM conditioning). All configurations are summarised in Table~\ref{tab:experiments}. We train with AdamW on the mean squared reconstruction error ($\approx 3.5$ hours on a single A100). Due to memory constraints, training uses a spatial subsample of $N \approx 780$ locations; all results are computed on the full global grid over the held-out test period.

\begin{table}[h]
  \centering
  \caption{Experiment configurations.}
  \label{tab:experiments}
  \small
  \begin{tabular}{@{}llll@{}}
    \toprule
    Experiment                          & Surface conditioning         & Column chl-a & Surface chl-a \\
    \midrule
    Emulator baseline (original BG4Sea) & physics only                 & BIORYS4      & ---           \\
    Analysis (contribution)             & physics + chl-a              & BIORYS4      & GlobColour    \\
    \midrule
    Oracle (diagnostic)                 & physics + chl-a              & SOCA         & SOCA          \\
    Ablation (diagnostic)               & none (column-only, no FiLM)  & SOCA         & ---           \\
    \bottomrule
  \end{tabular}
\end{table}

\section{Results}

Table~\ref{tab:results} reports skill against SOCA for the configurations in Table~\ref{tab:experiments}, with BIORYS4 included as reference. We use both pointwise metrics (RMSE, $R^{2}$, Pearson $r$) and distributional ones: Kolmogorov--Smirnov distance (KS), Cramér--von Mises statistic (CvM), and root mean squared quantile bias (QQ bias).

\begin{table}[h]
  \centering
  \caption{Skill against SOCA for the configurations in Table~\ref{tab:experiments}; BIORYS4 is the emulated simulation itself. Arrows give the direction of better skill. Bold marks the best value per metric, excluding the oracle.}
  \label{tab:results}
  \small
  \begin{tabular}{@{}lrrrrrr@{}}
    \toprule
                      & RMSE $\downarrow$ & $R^{2}$ $\uparrow$ & $r$ $\uparrow$  & KS $\downarrow$ & CvM $\downarrow$ & QQ bias $\downarrow$ \\
    \midrule
    BIORYS4           & 0.1648            & 0.2899             & 0.6517          & 0.1848          & 1,133,378        & 0.0550               \\
    Emulator baseline & 0.1647            & 0.2908             & 0.6540          & 0.1878          & 1,155,096        & \textbf{0.0545}      \\
    Analysis          & \textbf{0.1222}   & \textbf{0.6097}    & 0.7811          & \textbf{0.1401} & \textbf{421,162} & 0.0918               \\
    \midrule
    Oracle            & 0.0799            & 0.8333             & 0.9154          & 0.0670          & 149,477          & 0.0251               \\
    Ablation          & 0.1244            & 0.5955             & \textbf{0.8018} & 0.1732          & 986,609          & 0.0671               \\
    \bottomrule
  \end{tabular}
\end{table}

The emulator baseline performs almost identically to BIORYS4, and satellite conditioning substantially improves the results: RMSE drops from 0.165 to 0.122, $R^{2}$ doubles to 0.61, and $r$ rises from 0.65 to 0.78. Distributional statistics show reductions too, with the exception of QQ bias, which rises from 0.055 to 0.092. This is at least in part attributable to a covariate shift between GlobColour and SOCA. In fact, by most metrics, the analysis agrees more closely with SOCA at the surface than GlobColour itself does (RMSE 0.138 vs.\ 0.147; $R^{2}$ 0.69 vs.\ 0.45; $r$ 0.84 vs.\ 0.80; KS 0.165 vs.\ 0.163; CvM 4,727 vs.\ 27,269; QQ bias 0.143 vs.\ 0.138, not shown in Table~\ref{tab:results}), indicating the model partially corrects the injected observation. A depth, quantile-resolved breakdown of QQ bias is provided in the Appendix (Figure~\ref{fig:qq_gc}).

These results also show that conditioning on satellite surface matches or beats injecting the validation target itself directly into the column. The table shows that the column-only ablation (no surface conditioning) improves the baseline, but adding FiLM surface conditioning pushes RMSE from 0.12 to 0.08 and $R^{2}$ from 0.6 to 0.83, all while substantially reducing all distributional metrics. This suggests that the surface conditioning carries most of the assimilation-relevant information.

\paragraph{Improvements across depth}

\begin{figure}[h]
  \centering
  \includegraphics[width=0.8\textwidth]{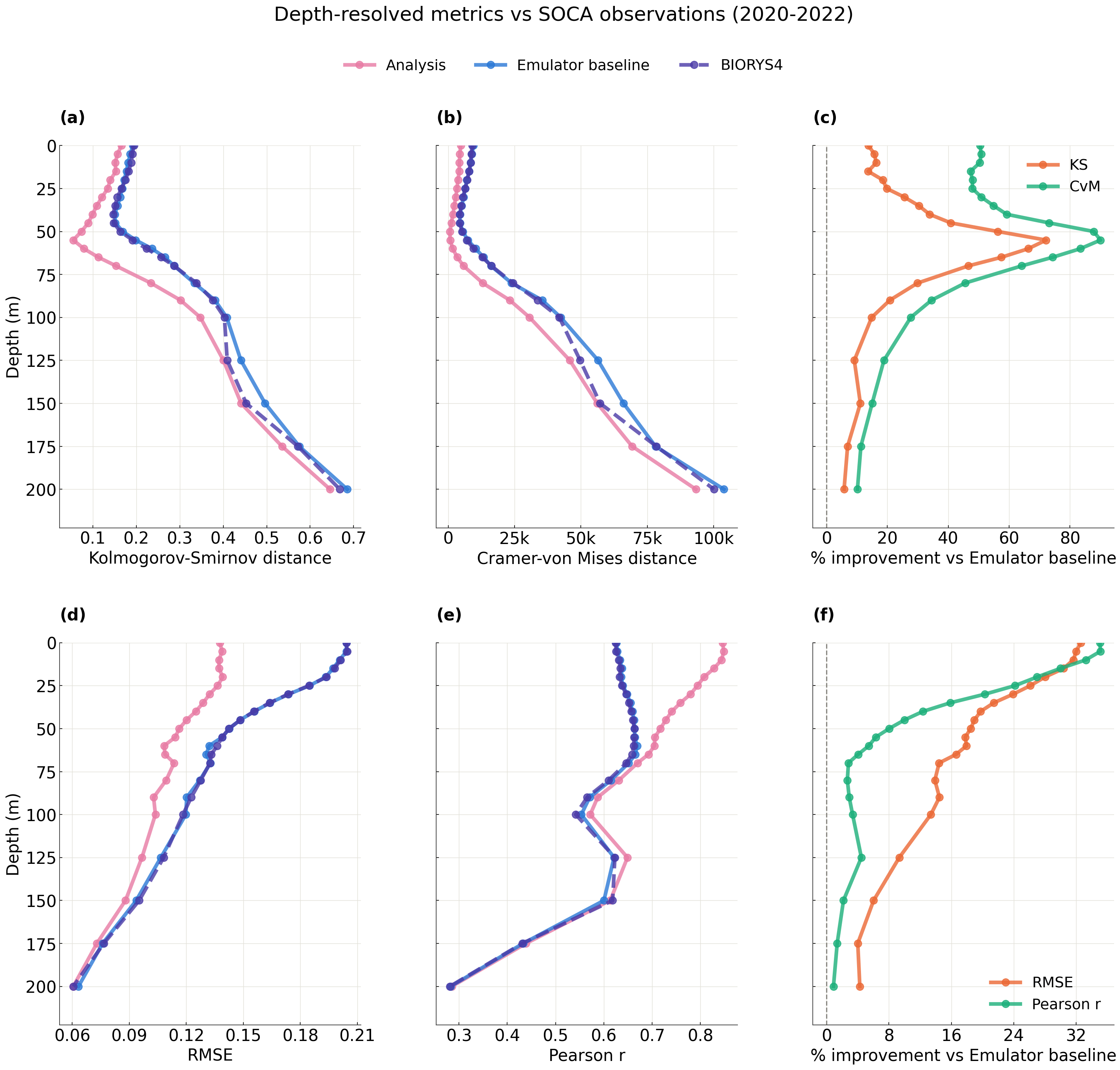}
\caption{Depth-resolved skill against SOCA for the analysis, emulator baseline, and BIORYS4: (a) KS, (b) CvM, (d) RMSE, (e) Pearson $r$. Panels (c) and (f) show the analysis' relative improvement over the baseline (positive is better).}
  \label{fig:met_depth}
\end{figure}

Figure~\ref{fig:met_depth} shows the depth-resolved skill. Although only a single surface scalar is changed, the analysis improves on the baseline and BIORYS4 at all depths and for all four metrics. Pointwise gains (RMSE, $r$) are largest at the surface (${\sim}33\%$) and decay with depth, whereas distributional gains (KS, CvM) peak in the subsurface at around 50--60\,m (70--90\%), before decaying smoothly below. This is also where the QQ bias reaches its minimum (Appendix Figure~\ref{fig:qq_gc}). 

\paragraph{Multivariate, trophically-consistent response}

To see how the single-variable correction spreads to the other outputs, we measure, for each variable $v$, how far the analysis moves from the simulated prior relative to the emulator's own reconstruction error, $\eta_v = \mathrm{RMSE}(\hat{\mathbf{x}}^{a}_v, \mathbf{x}_v) / \mathrm{RMSE}(\hat{\mathbf{x}}_v, \mathbf{x}_v)$, where $\mathbf{x}_v$ is BIORYS4. A ratio $\eta_v \approx 1$ indicates little to no departure from the prior, while larger values indicate a response. The ordering (Table~\ref{tab:tracer_response}) is qualitatively consistent with what one would expect from assimilating total chl-a into a BGC model: the chl-a pools and phytoplankton biomass respond most strongly, followed by primary production, grazers (zooplankton), and the nutrients that limit plankton growth. This is not an artefact of variables being harder to reconstruct, as $\eta_v$ is uncorrelated with per-variable reconstruction $R^{2}$ (Spearman correlation $-0.095$).

\begin{table}[htbp]
  \centering
  \caption{Response ratio $\eta_v$ (analysis-to-BIORYS4 over baseline-to-BIORYS4 RMSE) of each output variable to the surface chl-a substitution. DIC/DOC: dissolved inorganic/organic carbon; POC: particulate organic carbon; PAR: photosynthetically active radiation.}
  \label{tab:tracer_response}
  \small
  \begin{tabularx}{\linewidth}{@{}lX@{}}
    \toprule
    $\eta_v$             & Variables                                                                                                   \\
    \midrule
    $>2.5\times$         & Chl-a of diatoms and nanophytoplankton; total phytoplankton and diatom biomass                              \\
    $1.5$--$2.5\times$   & Primary production; meso\-zooplankton and nanophytoplankton biomass; Fe and Si content of phytoplankton      \\
    $1.1$--$1.5\times$   & NO$_3$, PO$_4$, O$_2$, DOC, CaCO$_3$, PAR, pH, large POC, microzooplankton biomass, salinity                 \\
    $\leq 1.1\times$     & DIC, NH$_4$, alkalinity, Fe, Si, temperature; sinking pools (biogenic Si, small POC, particulate Fe)        \\
    \bottomrule
  \end{tabularx}
\end{table}

Temperature and salinity are both column outputs and, at the surface, part of the physical conditioning $\mathbf{s}_{\mathrm{phys}}$, so they share the same MLP embedding as the substituted chl-a. In BIORYS4 the physics is prescribed and unaffected by the BGC, so ideally they should not respond at all. Their response is limited, negligible for temperature ($1.02\times$) and modest for salinity ($1.14\times$), below most biologically coupled tracers. This likely reflects water-mass correlations learned from the simulation rather than a physical pathway, suggesting the correction injects little spurious variance into the physical variables, although salinity's response deserves closer investigation. 

\section{Discussion}

Our results suggest that the adapted autoencoder acts as a simplified analysis operator, and that this is essentially zero-shot, sim-to-real transfer: the model, trained only on BIORYS4, maps GlobColour through the structure it learned, propagating a single surface scalar into a depth-resolved correction with no explicit gain matrix, covariance model, or optimisation. The substitution improves agreement with the float-based observational product at all depths, and is comparable to a much stronger intervention: the surface conditioning alone (RMSE 0.122, $R^{2}$ 0.61) matches replacing the chl-a column with the validation target itself in an unmodulated (column-only) network (RMSE 0.124, $R^{2}$ 0.60). The response follows a pattern that is qualitatively consistent with trophic relationships; the chl-a pools and phytoplankton biomass respond most strongly, and primary production and grazers follow. This reach likely comes from the intervention's modulation of the entire decoder and computing an explicit sensitivity map, as well as characterising how the conditioning embedding itself shifts when the observation is swapped, would be interesting directions for future work. Quantitatively validating the multivariate response is also an open direction. 

Interestingly, the network also handles the covariate shift between the two observational datasets gracefully. After conditioning, the analysis agrees with SOCA more closely than GlobColour does on RMSE, $R^{2}$, $r$ and CvM, in spite of GlobColour being its only source of information about the observed state. The decoder does not pass the observation unchanged, but it still inherits some of its biases (e.g. the satellite's narrower dynamic range, see Appendix Figure~\ref{fig:qq_gc}) closest to the injection point. Rescaling the conditioning to match SOCA would probably help, but the two products are different by construction, and that difference is what makes validation against SOCA relevant. Probabilistic conditioning could modulate how strongly the injected signal overrides the prior. The partial correction suggests that the structure provides some of this regulation implicitly, but with our deterministic approach it cannot be controlled.

\subsection{Limitations}

Perhaps the most important limitation of this work is that everything is done deterministically, and so we end up with a single point estimate with no notion of uncertainty arising from the distributional differences between the prior and the two observational distributions. The analysis inherits GlobColour's compressed upper tail (Appendix Figure~\ref{fig:qq_gc}), and there is no mechanism to hedge against this bias.

Our results examine the trophic response with respect to BIORYS4, but for variables other than chl-a there is no independent validation against observations. The correction propagates through correlations learned from the simulation rather than causal pathways, and it is unknown whether these reflect the correct covariance structure. While the appeal of our approach lies in its cost (single forward pass rather than an iterative analysis), we do not compare against conventional data assimilation, which we considered beyond the scope of this proof of concept. 

Furthermore, while the full emulator architecture incorporates spatial context via neighbourhood cross-attention in the latent forecaster~\cite{BG4Sea}, the decoder is strictly 1D, and each column is processed in isolation. Integrating horizontal structures remains a prerequisite for a full 3D correction. Finally, even though this is intended for forecast correction, evaluation is limited to autoencoder reconstruction. This relies on two assumptions: that a perfect column forecast coincides with the autoencoder's latent representation, and that a perfect forecaster exists to propagate the conditioning observation to the target time of the forecast. Both remain to be validated.

\section{Conclusion}

We show that the conditioning pathway of an ocean BGC emulator, trained only on simulation output, can be adapted to serve as a zero-shot analysis step. Substituting chl-a at the surface with its observation matches the replacement of the full chl-a column in the unconditioned network, improving agreement with the validation observations at all depths, and agreeing with them more closely than the injected observation itself. The correction propagates across variables in a pattern qualitatively consistent with trophic relationships. The main drawback is the lack of uncertainty quantification inherent to the deterministic formulation, making probabilistic extensions the natural next step towards a more robust analysis.

\bibliographystyle{unsrtnat}
\bibliography{bibliography}


\newpage
\appendix
\section{Examining the increased QQ bias}

The QQ bias in Table~\ref{tab:results} worsens under GlobColour conditioning (0.055 to 0.092). This section examines where and why.

\begin{figure}[h]
  \centering
  \includegraphics[width=\textwidth]{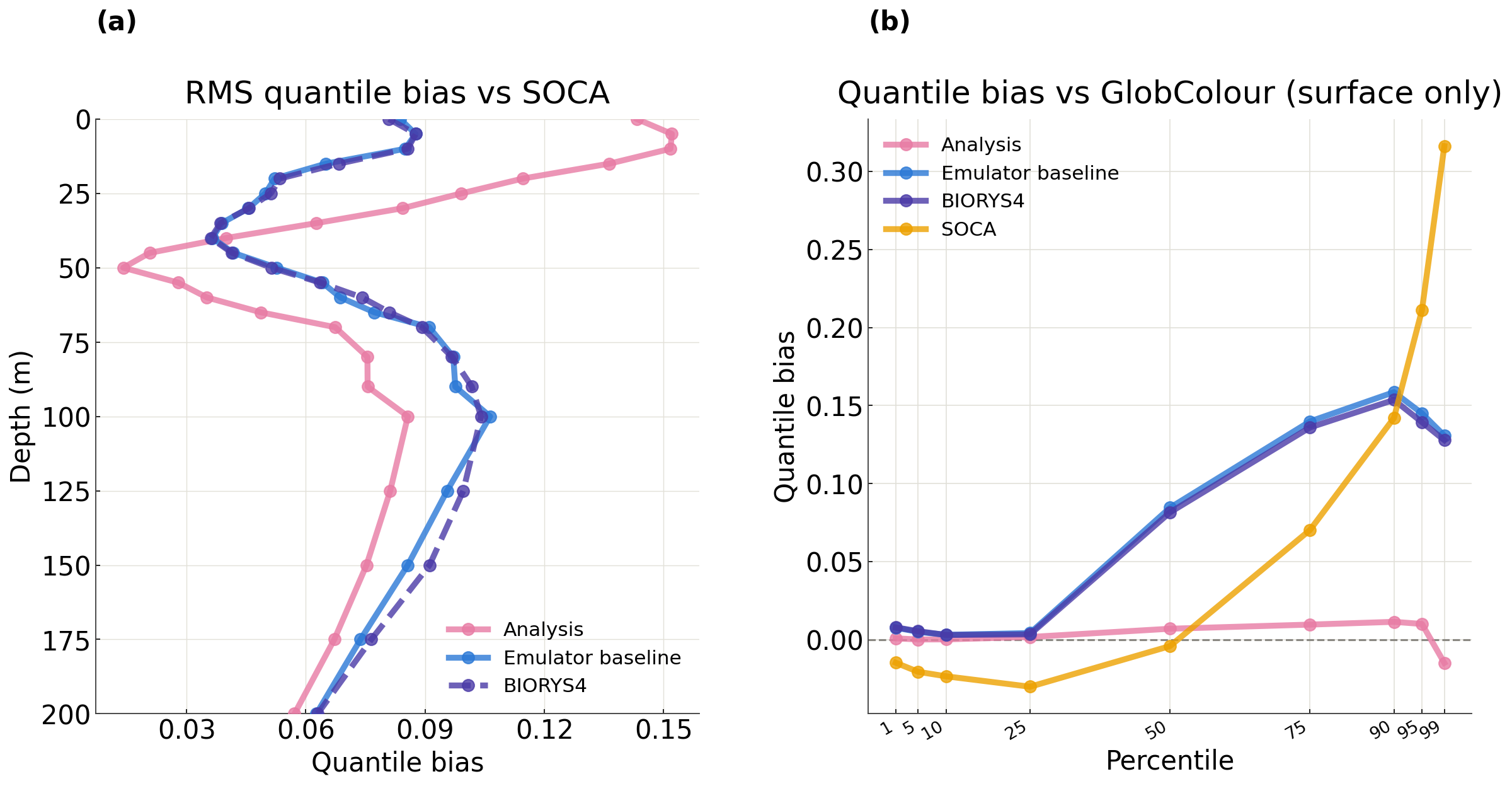}
  \caption{QQ bias breakdown. Panel A: Depth-resolved RMS QQ bias shows the increase concentrated near the surface. Panel B: Surface quantiles reveal that the analysis tracks GlobColour closely, while SOCA diverges at high percentiles.}
  \label{fig:qq_gc}
\end{figure}

Panel A of Figure~\ref{fig:qq_gc} shows that the bias increases at the surface, peaking at around 10m and rapidly decreasing to a minimum at around 50m. Below this depth, the analysis improves the metric relative to the baselines, showing that the bias is confined near the top.

Panel B illustrates the bias at the surface with respect to GlobColour (the injected observation), showing the differences by quantile. Here we see the analysis closely aligning with the source, while SOCA diverges sharply at high percentiles, indicating that GlobColour systematically underestimates peaks relative to the in situ measurements. This is consistent with what the literature describes: the satellite observations consistently show lower dynamic range~\cite{huChlorophyllAlgorithmsOligotrophic2012, baudenaComparingSatelliteBGCArgo2025, johnsonThreeImprovedSatellite2013, garnesson_cmems_2019-1}.

\section{Full list of variables and depths}

\begin{table}[htbp]
  \centering
  \caption{Depth-column variables (31), grouped by process.}
  \label{tab:all_vars}
  \begin{tabular}{@{}cccc@{}}
    \toprule
    \begin{minipage}[t]{0.22\textwidth}\raggedright
      \textbf{Nutrients and oxygen}\\[0.35em]
      Nitrate\\
      Phosphate\\
      Ammonium\\
      Total dissolved iron\\
      Diatom iron content\\
      Nanophytoplankton iron content\\
      Large particulate iron \\
      Small particulate iron \\
      Silicate\\
      Diatom silicon content\\
      Sinking biogenic silica \\
      Oxygen
    \end{minipage}
     &
    \begin{minipage}[t]{0.22\textwidth}\raggedright
      \textbf{Biology}\\[0.35em]
      Nanophytoplankton biomass\\
      Diatom biomass\\
      Total phytoplankton biomass\\
      Total chlorophyll\\
      Nanophytoplankton chlorophyll\\
      Diatom chlorophyll\\
      Net primary production\\
      Microzooplankton biomass\\
      Mesozooplankton biomass
    \end{minipage}
     &
    \begin{minipage}[t]{0.20\textwidth}\raggedright
      \textbf{Carbon pools}\\[0.35em]
      Alkalinity\\
      Dissolved inorganic carbon\\
      Dissolved organic carbon\\
      Calcium carbonate (calcite)\\
      pH \\
      Small particulate organic carbon\\
      Large particulate organic carbon
    \end{minipage}
     &
    \begin{minipage}[t]{0.20\textwidth}\raggedright
      \textbf{Physics \& light}\\[0.35em]
      Temperature\\
      Salinity\\
      Photosynthetically active radiation
    \end{minipage}
    \\
    \bottomrule
  \end{tabular}
  \medskip
\end{table}

\begin{table}[htbp]
  \centering
  \begin{subtable}[t]{0.48\textwidth}
    \centering
    \caption{Nominal depth levels (24) on the model vertical grid.}
    \label{tab:depths}
    \small
    \begin{tabular}{@{}rrrrrr@{}}
      \toprule
      \multicolumn{6}{c}{\textbf{Depth (m)}}       \\
      \midrule
      0.5  & 1.5   & 3.8   & 5.1   & 7.9   & 11.4  \\
      13.5 & 15.8  & 18.5  & 21.6  & 25.2  & 29.4  \\
      34.4 & 40.3  & 47.4  & 55.8  & 65.8  & 77.9  \\
      92.3 & 109.7 & 130.7 & 155.9 & 186.1 & 222.5 \\
      \midrule
    \end{tabular}
    \medskip
  \end{subtable}
  \hfill
  \begin{subtable}[t]{0.48\textwidth}
    \centering
    \caption{Surface variables used for conditioning}
    \label{tab:surface_cond}
    \small
    \begin{tabular}{@{}l@{}}
      \toprule
      \textbf{Surface-only conditioning} \\
      \midrule
      \textbf{Chlorophyll-a }            \\
      Mixed-layer depth                  \\
      Sea-surface height                 \\
      Sea-surface salinity               \\
      Sea-surface temperature            \\
      Particulate dust                   \\
      \bottomrule
    \end{tabular}
  \end{subtable}
  \caption{Depth grid and surface inputs}
\end{table}

\end{document}